\pdfoutput=1
\documentclass[11pt]{article}

\usepackage[final]{acl}
\usepackage{times}
\usepackage{latexsym}

\usepackage[T1]{fontenc}
\usepackage[utf8]{inputenc}

\usepackage{microtype}

\usepackage{inconsolata}

\usepackage{graphicx}
\usepackage{multirow}
\usepackage{CJKutf8}
\usepackage{amssymb}

\title{Fostering Natural Conversation in Large Language Models with NICO: a Natural Interactive COnversation dataset}

\author{
    Renliang Sun$^{\dag\ddag}$\thanks{Equal contribution} \quad
    Mengyuan Liu$^{\dag\S}$\textsuperscript{*} \quad
    Shiping Yang$^{\P}$\textsuperscript{*}
    \\
\hspace{8mm}\textbf{
    Rui Wang$^{\dag}$ \quad
    Junqing He$^{\dag}$ \quad
    Jiaxing Zhang$^{\dag}$
    }
    \\
\hspace{2mm}
    $^\dag$International Digital Economy Academy \quad $^\ddag$University of California, Los Angeles \quad \\
    $^\S$University of Michigan, Ann Arbor \quad 
    $^\P$Simon Fraser University\\
    \hspace{2mm}{\tt sunrenliang@ucla.edu, juneliu@umich.edu, yangshipingnlp@gmail.com} \\
    {\tt \{wangrui, hejunqing, zhangjiaxing\}@idea.edu.cn} 
}

\begin{document}
\maketitle
\begin{abstract}

Benefiting from diverse instruction datasets, contemporary Large Language Models (LLMs) perform effectively as AI assistants in collaborating with humans. However, LLMs still struggle to generate natural and colloquial responses in real-world applications such as chatbots and psychological counseling that require more human-like interactions. To address these limitations, we introduce NICO, a Natural Interactive COnversation dataset in Chinese. We first use GPT-4-turbo to generate dialogue drafts and make them cover 20 daily-life topics and 5 types of social interactions. Then, we hire workers to revise these dialogues to ensure that they are free of grammatical errors and unnatural utterances. We define two dialogue-level natural conversation tasks and two sentence-level tasks for identifying and rewriting unnatural sentences. Multiple open-source and closed-source LLMs are tested and analyzed in detail. The experimental results highlight the challenge of the tasks and demonstrate how NICO can help foster the natural dialogue capabilities of LLMs. The dataset will be released.

%The results of the human evaluation illustrate that NICO is consistent with human speaking habits and high-quality.

\end{abstract}

\section{Introduction}
The development of large language models (LLMs) has received much attention over the past few years, demonstrating their potent potential across numerous fields \cite{zhao2023survey}. In particular, in the field of dialogue, these models have been widely used in areas such as chatbots \cite{zheng2023judging} and virtual assistants \cite{liu2023dynamic}.

%这一段我感觉没有说明白之前研究的不足在哪，没有突出我们研究的目的和现实意义
%目前的SFT dataset更关注提升LLM各方面的能力，让LLM扮演一个更强大的助手，但是忽略了如chatbot、心理支持等场景中需要让LLM表现的更Human-like的需求。当前的LLM在进行对话任务时仍然饱受不够自然及不够口语化等问题的困扰。先前的研究表明即使是GPT4也只有百分之五十的情况能通过图灵测试，由于...等原因。
%to do: english version
Many researchers have sought to enhance the performance of LLMs by designing instruction datasets. Existing datasets for fine-tuning LLMs can be categorized into those that improve instruction-following abilities \cite{he2024can}, planning abilities \cite{dagan2023dynamic}, reasoning abilities \cite{wang2023can}, etc. These datasets have indeed improved the performance of LLMs on benchmarks such as MMLU \cite{hendrycks2020measuring} and MMLU-Pro \cite{wang2024mmlu}. However, when engaging in dialogue with humans, the responses generated by LLMs are often unnatural and insufficiently colloquial \cite{guo2023evaluating,yi2024survey}. LLMs can only act as AI assistants rather than human beings. % The models often struggle to understand the context of the dialogue and generate reasonable and colloquial responses \cite{guo2023evaluating,yi2024survey}. 
It has led to the failure of the LLMs for applications such as psychological counseling and accompanying children. \citet{jones2024does} showed that the success rate of the best-performing GPT-4 in acting as a human was only 50\%. The main reasons for the failure included unnatural and too informal responses, weird tones, and so on.

To address these issues, we are committed to constructing a multi-turn daily dialogue dataset in Chinese. We aim to evaluate the ability of LLMs to simulate natural human conversations in daily life, and improve their abilities by fine-tuning with the constructed dataset.

%data construction
We ensure that the dialogues in this dataset are related to daily life and conform to human speaking habits. We first define 20 broad topics, which are basically involved and discussed in human daily life. For each topic, we construct 200 specific concepts or items. We also draw inspiration from the five types of interactions in sociology. To make the constructed dialogs resemble real human conversations, we ensure that the dataset covers all interaction types. Then, we utilize GPT-4-turbo to generate 4,000 dialogs, which contains 22.1 utterances per dialog on average. We name the constructed dataset NICO. We also perform a human evaluation, and the results show that NICO is better than the current representative datasets in fluency, coherence and naturalness.

%How to use NICO
Subsequently, we define four tasks, two of which are used to test the daily conversation capabilities of LLMs. The other two are used to test the capabilities to identify and rewrite unnatural sentences.
We have selected several representative LLMs for testing and also conducted supervised fine-tuning. The results indicate that mimicking human-like daily conversation poses a challenge for LLMs. The dialogs we constructed can enhance the performance of LLMs. 
At the sentence level, LLMs can rewrite unnatural sentences but are almost incapable of identifying them. Further analysis of the results reveals that LLMs perform much worse in conflict-type dialogues than in other types of dialogues, with limited improvement from fine-tuning.

In summary, enabling LLMs to mimic human-like fluent and natural daily conversations remains a challenge. The contributions of our paper are as follows:

\begin{itemize}
\item We construct a multi-turn natural interactive conversation dataset in Chinese named NICO, covering 20 daily topics and 5 interaction types. Human annotation ensures that NICO is free of grammatical errors and unnatural utterances.
% The evaluation results show that the quality of NICO surpasses existing datasets.
\item We define four tasks, including two dialogue-level tasks to test the natural dialog abilities of LLMs and two sentence-level tasks of identifying and rewriting unnatural sentences. We test several representative LLMs on the four tasks and report their results.
%最后一句太短了
%We evaluate several representative LLMs on these four tasks and report the results, which can serve as a reference for future research.
\item We conduct a detailed analysis of the results, pointing out the deficiencies of existing LLMs in conducting natural conversations and identifying unnatural sentences. We validate that NICO can help LLMs to generate responses that are more human-like rather than resembling AI assistants.
% NICO can help improve the natural dialog abilities of LLMs and provide a reference for future research.
%our data ... provide a reference 怪怪的
%Through tailored ablation experiments, we validate the effectiveness of NICO in imporving the natural dialogue abilities of LLMs.
% 像人而不是助手
\end{itemize}

\section{Related Work}

In recent years, open-domain dialog systems have witnessed tremendous progress as the number of model parameters keeps growing. Representative dialog systems include ChatGPT \cite{achiam2023gpt}, Gemini \cite{reid2024gemini}, Mixtral-Instruct \cite{jiang2024mixtral}, LLama3-Instruct \cite{meta2024introducing}, and so on \cite{cai2024internlm2, gan2023ziya2,wei2023skywork,yang2023baichuan}. 
% These models are usually pre-trained on a large amount of knowledge-based texts such as books and newspapers. Then, they undergo supervised fine-tuning (SFT) and reinforcement learning from human feedback (RLHF) on a small amount of dialog data and instruction data. 

Many scholars have been working on constructing dialog data to help dialog systems evolve. Some datasets focus on human emotions. DailyDialog \cite{li2017dailydialog} is an English conversation dataset containing various daily conversation scenarios. % Its conversations cover a wide range of topics in daily life. 
% This dataset especially emphasizes the communication intention and emotion information of the conversations.
% Persona-chat \cite{zhang2018personalizing} is an English conversation dataset that focuses on daily communication. It makes conversations more like human-to-human communication by establishing background information about different users. 
Persona-chat \cite{zhang2018personalizing} is an English conversation dataset that makes conversations like human-to-human communication by establishing users' background information.
%这段过渡的不是很明确，strategy后面好像没提到
Other datasets focus on strategy and behavior. The Ubuntu Dialogue Corpus \cite{lowe2015ubuntu} is a multi-turn dialog dataset in English collected from technical support conversations. MultiWOZ \cite{budzianowski2018multiwoz} is also a large-scale English conversation dataset. Each of its conversations is labeled with a sequence of states and the respective system dialog behavior. % There are also many other high quality English dialog corpus.
Sotopia \cite{zhousotopia} and Sotopia-$\pi$ \cite{wang2024sotopia} involve methods to generate dialog data and enhance LLM's ability to accomplish social goals.

Apart from the above English conversation datasets, many researchers working on proposing high-quality Chinese conversation datasets and developing Chinese conversation systems. Some Chinese conversation datasets such as Douban Conversation Corpus \cite{wu2017sequential}, LCCC \cite{wang2020large}, and CConv \cite{li2022c3kg} are mined and cleaned from Chinese social media. Some datasets such as E-commerce Dialogue Corpus \cite{zhang2018modeling}, JDDC corpus \cite{chen2020jddc} are constructed from e-commerce service stuff and customer conversation data. There are also datasets such as DuConv \cite{wu2019proactive}, Mutual \cite{cui2020mutual}, KdConv \cite{zhou2020kdconv}, Corecode \cite{shi2024corecode} annotated using knowledge bases including news and exams.

Despite the above efforts, existing dialog systems are still considered to be inconsistent with human speaking habits \cite{jones2024does}. 
Natural dialogue datasets, especially those in Chinese, are still scarce. High-quality Chinese dialog datasets are essential to advance the progress of Chinese dialog systems.

\section{The NICO Dataset}
%我在想这个部分要不要简略说明一下我们选择用gpt4 annotation + 人工的原因以及好处，再引用一些之前的工作来support我们这样构建数据集的合理性。
%模型提供对话框架及相关背景知识，人为矫正语法错误以及使其口语化。大幅降低了对话标注的难度以及成本，并且还提高了标注的质量。（允许我们将标注的范围缩小到一部分高水平标注者当中）
%to do: english version
We have built a Natural Interactive COnversation dataset named NICO based on daily topics and the social interactions. In this section, we will detail how we build and annotate the dialogues, as well as ensure the quality of the data.

\subsection{Data Construction}

We aim to construct dialogues that are as relevant as possible to real life. We start by defining 20 broad topics, including culture, technology, and so on. These topics are fundamentally ones that humans frequently engage with and discuss in their daily lives. The complete broad topics are shown in Table \ref{tab:topics}. Then, we give prompts for GPT-4-turbo \cite{achiam2023gpt} to generate 200 specific items or concepts under each broad concept, e.g., sweaters, artificial intelligence, etc. We carefully check and modify these generations to ensure their alignment with the broad topics. 
% Please refer to Appendix \ref{sec:appendix_topics} for the complete broad topics.

\begin{table}[ht!]
\centering
\resizebox{\columnwidth}{!}{%
\begin{tabular}{l|l}
\hline
Topics &
  \begin{tabular}[c]{@{}l@{}}\begin{CJK*}{UTF8}{gbsn}动物\end{CJK*}(Animal), \begin{CJK*}{UTF8}{gbsn}艺术和手工艺品\end{CJK*}(Art and Craft Supplies),\\ \begin{CJK*}{UTF8}{gbsn}图书和文具\end{CJK*}(Books and Stationery), \begin{CJK*}{UTF8}{gbsn}衣服\end{CJK*}(Clothing), \begin{CJK*}{UTF8}{gbsn}文\end{CJK*}\\ \begin{CJK*}{UTF8}{gbsn}化\end{CJK*}(Culture), \begin{CJK*}{UTF8}{gbsn}教育和学习\end{CJK*}(Education and Learning), \begin{CJK*}{UTF8}{gbsn}金融\end{CJK*}\\ \begin{CJK*}{UTF8}{gbsn}产品和服务\end{CJK*}(Financial Products and Services), \begin{CJK*}{UTF8}{gbsn}饮食\end{CJK*}\\ (Food and Drinking), \begin{CJK*}{UTF8}{gbsn}杂货和家庭必需品\end{CJK*}(Groceries and \\ Household Essentials), \begin{CJK*}{UTF8}{gbsn}健康和美容\end{CJK*}(Health and Beauty),\\ \begin{CJK*}{UTF8}{gbsn}历史和事件\end{CJK*}(History and Events), \begin{CJK*}{UTF8}{gbsn}音乐和乐器\end{CJK*}(Music \\ and Instruments), \begin{CJK*}{UTF8}{gbsn}植物\end{CJK*}(Plant), \begin{CJK*}{UTF8}{gbsn}不动产\end{CJK*}(Real Estate), \begin{CJK*}{UTF8}{gbsn}体\end{CJK*}\\ \begin{CJK*}{UTF8}{gbsn}育运动\end{CJK*}(Sports), \begin{CJK*}{UTF8}{gbsn}科技\end{CJK*}(Technology), \begin{CJK*}{UTF8}{gbsn}玩具和游戏\end{CJK*}(Toys \\ and Games), \begin{CJK*}{UTF8}{gbsn}交通运输\end{CJK*}(Transportation), \begin{CJK*}{UTF8}{gbsn}旅行\end{CJK*}(Travel),\\ \begin{CJK*}{UTF8}{gbsn}工作和职业\end{CJK*}(Working and Occupation)\end{tabular} \\ \hline
\end{tabular}%
}
\caption{The 20 defined broad topics.}
\label{tab:topics}
\end{table}

In sociological terms, the process of communication between members of a society is known as social interaction. 
% Social interaction is a basic unit of analysis in sociology. 
It describes the way people behave when they cross paths with others. Social interaction is consistent with the definition of behavioral intelligence proposed by \citet{li2024social}. and has significant ramifications for human-AI interactions. We argue that high-quality daily dialog datasets need to cover all types of social interactions. There are many types of interactions between humans, and in the paper we accommodate the widely accepted types of social interactions proposed by \citet{nisbet1970social}.

\begin{quote}
\itshape
Five modes or types of social interaction are probably the most basic and universal ones: exchange, cooperation, conformity, coercion, and conflict.
\begin{flushright}
--- \cite{nisbet1970social}, p.63
\end{flushright}
\end{quote}

We construct dialogs based on the five interaction types and give the definition of each interaction type as follows:
\begin{itemize}
    \item \textbf{Exchange} refers to the process in which an individual or a group takes action with the explicit purpose of receiving a return. 
    \item \textbf{Cooperation} is a collaborative behavior to achieve a certain goal of common interest.
    \item \textbf{Conformity} is the process in which an individual's behavior adapts to a pre-existing norm, in which the influence of the social group is either evident or subtle.
    \item \textbf{Coercion} is a behavior that is produced by compulsion of any form, often accompanied by threats.
    \item \textbf{Conflict} refers to the process in which individuals struggle with each other to compete for a scarce, commonly desired return.
\end{itemize}

Then, we write three example dialogues for each interaction type. Benefiting from the powerful data annotation capability of LLMs \cite{llmsforannotation, li2024dalk}, we provide the definitions of the interaction types, the corresponding dialogues, and the requirements for generating colloquial dialogues as prompts to GPT-4-turbo\footnote{The version is gpt-4-turbo-2024-04-09.}, which is asked to generate daily dialogues about specific concepts. A prompt example is shown in Figure \ref{generation prompt}. For each interaction type, we generate 800 dialogs for a total of 4,000. These dialogs also entirely cover the 4,000 generated specific items and concepts. 
% It costs roughly \$120 to generate these dialogs.

\subsection{Annotation Process}

%associated with好像不太合适
%Despite the advanced instruction-following and text-generation capabilities of large language models (LLMs), they are still prone to issues such as hallucinations \cite{yang2023new} and formatting errors \cite{sui2024table}. Therefore, we have employed native Chinese-speaking workers to review and correct the data generated by the LLMs. This process involves three stages of annotation.
Despite the strong instruction-following and text-generating capabilities of the LLMs, they still struggle with problems such as hallucinations \cite{yang2023new} and formatting errors \cite{sui2024table, tong2024can}. Therefore, we have employed workers to check and correct the generated data. All workers are native Chinese speakers enrolled in a bachelor's degree or higher. They read the instructions carefully and performed trial annotations before the annotation started. We rejected some workers with poor quality annotations in advance.
% The annotators were paid much more than the minimum hourly wage in their countries. 
Then, we conduct three stages of annotation. 

In the first stage, we ask workers to check whether the generated dialogs match the corresponding interaction types. 
If discrepancies are found, such as agreement and cooperation in a conflict-type dialogue, the worker will revise the dialogue directly to ensure that it is consistent with the intended interaction type.
%If discrepancies are found—such as a situation where the two parties agree and cooperate in a dialogue that is supposed to depict conflict—the worker will directly revise the dialogue to ensure it conforms to the intended social type.
% If not, for instance, the two parties agree and cooperate in a conflicting dialog, the worker will directly modify the dialog to make it conform to the dialog type. 
% 改一下
In the second stage, workers annotate sentences in the dialog that contain grammatical errors, sentences that do not conform to human speech i.e., are unnatural, and blocks of sentences that do not convey information i.e., are meaningless. In the third stage, workers rewrite sentences that contain grammatical errors and unnatural sentences. For cost reasons, each dialog is assigned to one worker. During the annotation process, we check the quality of the annotations throughout the process. If there are many labels that are unreasonable, the workers will be asked to re-do the annotation until it passes. We find that meaningless sentences mainly appear at the end of coercion-type dialogs and are not numerous. Therefore, we delete them directly, which will not affect the content of the dialog. 

% We display the total number of annotated problematic sentences and dialogues in Table \ref{tab:annotation}.

After the annotation is completed, we obtain 4,000 natural dialogs related to daily life. We further partition these dialogs into training, validation, and test sets, with 3,600, 200, and 200 dialogs respectively, for subsequent experiments.

\section{Statistics}

In this section, we have performed statistics on the features of the NICO dataset. We have also conducted human evaluations and compared the NICO dataset with other conversation datasets.

\subsection{Dataset Statistics}

The statistics of the constructed dataset are shown in Table \ref{tab:statistics}. Our dataset comprises 4000 dialogues and 88,545 utterances. On average, there are 22.1 utterances per dialog, suggesting sufficient rounds of interaction between the dialog's parties. The average number of tokens per utterance is 24.7, indicating that the dialog conveys adequate information. We also count the annotations and find that more than one-third of the dialogs contain grammatical errors, and more than half of the dialogs contain unnatural utterances. This suggests that the dialogues generated by LLMs are not always reliable and require manual correction to enhance their quality.

\begin{table}[ht!]
\centering
\resizebox{7cm}{!}{%
\begin{tabular}{l|c}
\hline
\# Dialogs                           & 4,000  \\
\# Utterances                        & 88,545 \\
\# Topics                            & 20     \\ \hline
Avg. \# utterances per dialog        & 22.1   \\
Avg. \# tokens per dialog            & 546.5  \\
Avg. \# tokens per utterance         & 24.7   \\ \hline
\# Dialogs with grammatical errors   & 1,440  \\
\# Utterances with grammatical errors & 2,017  \\
\# Dialogs with unnatural utterances  & 2,050  \\
\# Unnatural utterances               & 2,872  \\
\# Dialogs with meaningless utterances & 418    \\
\# Meaningless utterance blocks       & 418    \\ \hline
\end{tabular}%
}
\caption{Statistics of the NICO dataset.}
\label{tab:statistics}
\end{table}

\subsection{Comparisons}

To evaluate the quality of the NICO dataset, we select several representative conversation datasets for comparison: (1) NaturalConv \cite{wang2021naturalconv}, (2) KdConv \cite{zhou2020kdconv}, (3) LCCC \cite{wang2020large}, (4) DailyDialog \cite{li2017dailydialog}. The first three datasets are high-quality Chinese dialog datasets. NaturalConv and KdConv have been manually checked, while LCCC has been automatically filtered using a set of rules. DailyDialog is a human-written English dialog dataset, so we use GPT-4-turbo for translation. We randomly selected 100 dialogs from each of the above datasets and the NICO dataset we built for evaluation.

We have also performed a human evaluation. Following \cite{zhou2020kdconv}, we choose fluency and coherence as evaluation metrics. However, to emphasize the importance of naturalness in conversation, we only use fluency to measure grammatical correctness, and use naturalness as a separate evaluation metric following \cite{wang2021naturalconv}. The definitions are as follows and more details are in Appendix \ref{sec:evaluation_details}.

\noindent
\textbf{Fluency} (rating scale is 0,1,2) is defined as whether the dialog contains syntax errors.

\noindent
\textbf{Coherence} (rating scale is 0,1,2) is defined as whether the sentences in a dialog are relevant to the context and logically correct.

\noindent
\textbf{Naturalness} (rating scale is 0,1,2) is defined as whether the dialog is natural and conforms to human speaking habits.

We hire three native Chinese-speaking workers for the evaluation. To ensure the quality of the evaluation, we train them before the evaluation.
% and the evaluation guidelines are in Appendix Z. 
The evaluation results are shown in Table \ref{tab:datasets}.

% \usepackage{graphicx}
\begin{table}[ht!]
\centering
\resizebox{7cm}{!}{%
\begin{tabular}{l|ccc}
\hline
Datasets    & Fluency    & Coherence    & Naturalness    \\ \hline
NaturalConv & 1.68 & 1.95 & 1.70 \\
KdConv      & 1.86 & 1.96 & 1.47 \\
LCCC        & 1.33 & 0.95 & 1.03 \\
DailyDialog & 1.88 & 1.90 & 1.44 \\ \hline
NICO        & 1.98 & 2.00 & 1.90 \\ \hline
\end{tabular}%
}
\caption{The human evaluation results of the NICO dataset versus other representative dialogue datasets.}
\label{tab:datasets}
\end{table}

Overall, the NICO dataset performs best on all three metrics, especially scoring well above other datasets in fluency and naturalness.
In the development and evaluation of dialog systems, using the NICO dataset may yield better results. LCCC without human filtering has the lowest scores, which suggests that human annotation is essential for constructing high-quality dialog datasets. DailyDialog has good fluency and coherence after translation, but the dialogs have obvious signs of machine translation. NaturalConv scores higher in naturalness than the other datasets, but still lower than NICO.

\begin{figure}[ht!]
\centering
\includegraphics[width=7cm]{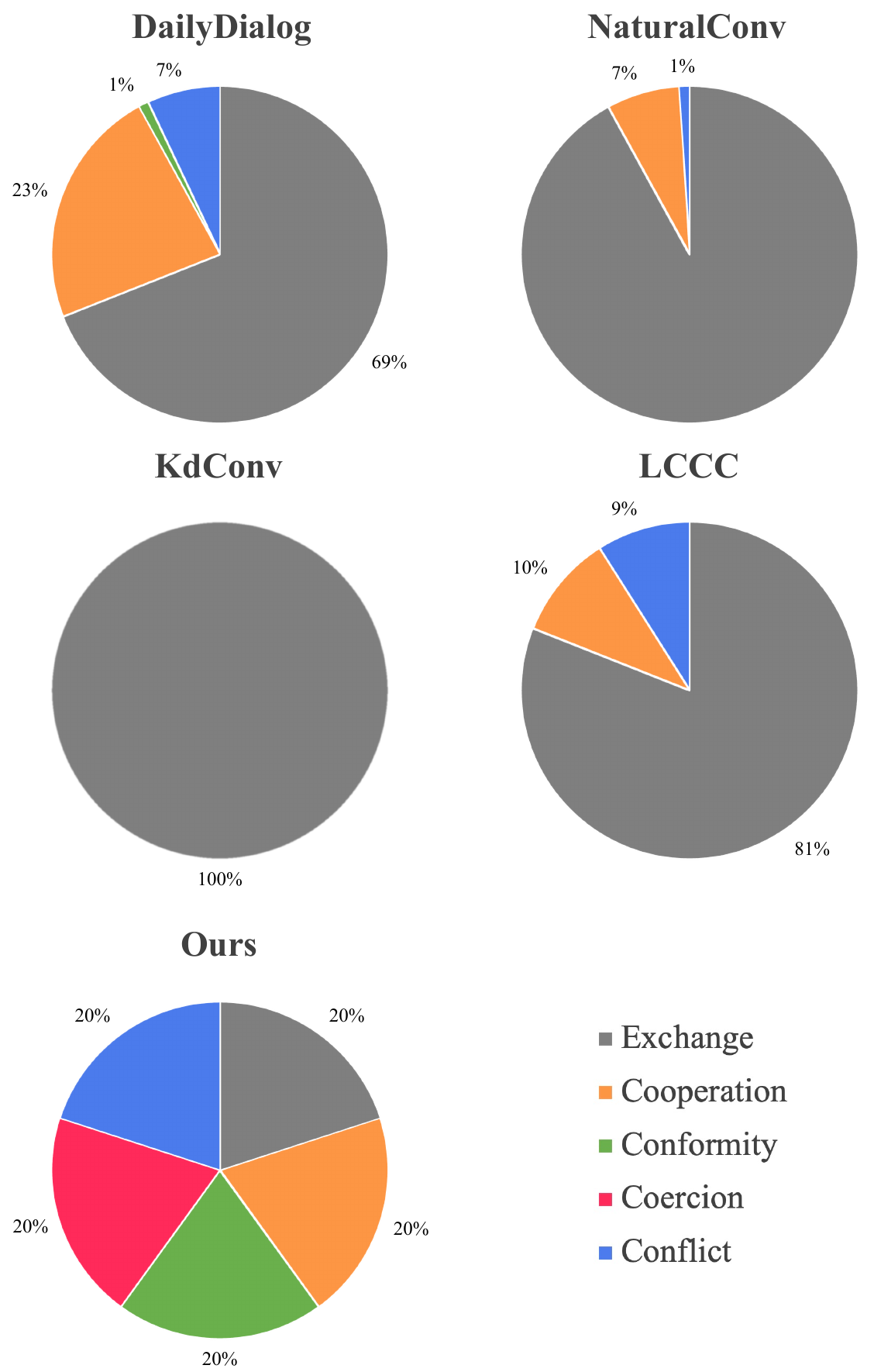} % Reduce the figure size so that it is slightly narrower than the column. Don't use precise values for figure width.This setup will avoid overfull boxes.
\caption{Distribution of interaction types of the dialog datasets. Only the constructed NICO dataset is able to cover all types of social interactions.}
\label{category}
\end{figure}
%说明一下我们进行这个比较的必要性
%为了说明引入交互类型的重要性，我们也对其余三个数据集中对话的社会交互类型进行了统计与比较。具体的说，我们手动为其余数据集创建了社会交互类型的标签
% 改一下，把详细分析加到caption里
We also perform statistics on the types of social interactions contained in these datasets. A high-quality dialog dataset should cover all five types of social interactions that occur in real, daily conversations. We let the workers choose the type of social interaction they believe is most closely represented in the dialogues. The results are depicted in Figure \ref{category}.

The most prevalent type of interaction across several comparison datasets is exchange. Both DailyDialog and LCCC contain a small number of cooperative and conflict dialogs. 
Almost none of the comparison datasets contain conformal dialog or coercive dialog. The constructed NICO dataset covers all social interaction types, thus aligning more closely with human communication patterns.

\section{Experiments}

\subsection{Tasks}

We have defined four tasks and tested LLMs on the constructed dataset. The first two tasks are designed to assess the abilities of LLMs to engage in natural dialogue in daily life scenarios. 
%这表述上应该需要提到剩下两个任务，不然有些疑惑
Benefiting from the process of annotating the dataset, we have the original dialog, the annotations of unnatural sentences in the dialog, and the corresponding human rewrites. We utilize these data for experiments on the last two tasks: identifying unnatural sentences and rewriting unnatural sentences.
% 可以加上我们是第一个这么做的。
% The last two tasks are aimed at testing the abilities to identify and rewrite unnatural sentences.

\noindent
\textbf{Open-domain dialog} This task is to let LLMs simulate a real human conversation. In this task, the user seems to be talking with a real person with multi-turn dialog history. It requires LLMs to have a broad knowledge base and strong contextual understanding ability, and to generate natural and fluent responses.
% This task is conducted in a multi-turn interaction format

\noindent
\textbf{Dialog completion} This task requires LLMs to generate a reasonable and natural response based on the instruction. 
%感觉描述的不够清楚
%Unlike the open-domain dialog task, the dialog history appears in instruction, and LLMs need to complete the dialog according to the given social interaction type. This task is conducted in a single-turn interaction format.
Unlike the open-domain dialog task, the multi-turn dialog appears in the instruction. The social interaction type with definition of the dialog is also given in the instruction.

\noindent
\textbf{Identifying unnatural sentences} This task involves enabling LLMs to identify sentences that do not conform to human speech patterns. The detected sentences can then be rewritten.

\noindent
\textbf{Rewriting unnatural sentences} This task is to have the LLMs rewrite sentences that do not conform to human speech patterns, making them more natural and fluent. It can enhance the readability of the text and the efficiency of human interaction.

Prompts for the above tasks are shown in Appendix \ref{sec:prompt_for_tasks}.

\begin{table*}[ht]
\centering
\resizebox{\textwidth}{!}{%
\begin{tabular}{lcccccccccc}
\hline
\multicolumn{1}{c|}{\multirow{2}{*}{Models}} &
  \multicolumn{5}{c|}{Open-domain Dialog} &
  \multicolumn{5}{c}{Dialog Completion} \\ \cline{2-11} 
\multicolumn{1}{c|}{} &
  BLEU-1/2 &
  ROUGE-L &
  Distinct-1/2 &
  $F_{\rm BERT}$ &
  \multicolumn{1}{c|}{Avg\_len} &
  BLEU-1/2 &
  ROUGE-L &
  Distinct-1/2 &
  $F_{\rm BERT}$ &
  Avg\_len \\ \hline
\multicolumn{11}{c}{Open-source LLMs} \\ \hline
\multicolumn{1}{l|}{Qwen2} &
  19.83/8.02 &
  23.29 &
  83.57/98.71 &
  64.30 &
  \multicolumn{1}{c|}{50.86} &
  17.78/6.55 &
  19.99 &
  84.72/99.33 &
  63.08 &
  56.00 \\
\multicolumn{1}{l|}{Qwen2(fs)} &
  24.66/10.54 &
  26.10 &
  89.32/99.55 &
  65.46 &
  \multicolumn{1}{c|}{33.31} &
  \underline{21.80}/8.02 &
  22.63 &
  {\underline{89.66/99.72}} &
  63.92 &
  36.49 \\
\multicolumn{1}{l|}{GLM-4} &
  13.82/5.87 &
  22.55 &
  70.79/94.03 &
  64.15 &
  \multicolumn{1}{c|}{77.16} &
  19.06/8.02 &
  21.98 &
  80.89/98.60 &
  64.41 &
  53.14 \\
\multicolumn{1}{l|}{GLM-4(fs)} &
  24.14/10.64 &
  26.20 &
  85.31/99.20 &
  65.86 &
  \multicolumn{1}{c|}{41.50} &
  20.90/\underline{8.42} &
  {\underline{22.94}} &
  83.61/99.12 &
  {\underline{64.64}} &
  44.76 \\
\multicolumn{1}{l|}{Yi-1.5} &
  24.38/10.71 &
  26.77 &
  86.53/98.78 &
  65.73 &
  \multicolumn{1}{c|}{34.43} &
  16.47/6.27 &
  19.16 &
  80.16/98.39 &
  62.87 &
  62.84 \\
\multicolumn{1}{l|}{Yi-1.5(fs)} &
  {\underline{26.25/12.33}} &
  {\underline{28.96}} &
  {\underline{91.71/99.47}} &
  {\underline{66.58}} &
  \multicolumn{1}{c|}{23.36} &
  18.04/6.52 &
  20.67 &
  81.56/98.59 &
  63.23 &
  54.66 \\
\multicolumn{1}{l|}{LLaMA3} &
  23.17/9.76 &
  26.29 &
  86.12/98.79 &
  65.10 &
  \multicolumn{1}{c|}{34.02} &
  18.79/6.94 &
  21.38 &
  81.46/98.69 &
  63.68 &
  55.48 \\
\multicolumn{1}{l|}{LLaMA3(fs)} &
  25.84/11.65 &
  28.90 &
  91.09/99.23 &
  66.45 &
  \multicolumn{1}{c|}{23.86} &
  17.90/5.98 &
  20.05 &
  83.02/98.76 &
  62.50 &
  52.56 \\ \hline
\multicolumn{11}{c}{Closed-source LLMs} \\ \hline
\multicolumn{1}{l|}{Doubao} &
  24.66/10.72 &
  27.30 &
  90.27/99.04 &
  66.20 &
  \multicolumn{1}{c|}{25.88} &
  24.28/10.18 &
  25.54 &
  88.86/99.33 &
  66.02 &
  29.92 \\
\multicolumn{1}{l|}{Doubao(fs)} &
  26.28/12.09 &
  29.28 &
  91.56/99.19 &
  66.83 &
  \multicolumn{1}{c|}{19.73} &
  26.78/11.94 &
  28.76 &
  \textbf{93.15/99.72} &
  67.06 &
  19.98 \\
\multicolumn{1}{l|}{Deepseek} &
  26.73/12.29 &
  28.39 &
  89.31/99.51 &
  66.60 &
  \multicolumn{1}{c|}{32.24} &
  18.00/7.58 &
  21.64 &
  80.76/98.81 &
  64.30 &
  65.91 \\
\multicolumn{1}{l|}{Deepseek(fs)} &
  29.22/14.92 &
  31.44 &
  \textbf{92.28/99.65} &
  67.92 &
  \multicolumn{1}{c|}{23.74} &
  24.03/10.75 &
  25.92 &
  86.07/99.37 &
  65.96 &
  41.38 \\
\multicolumn{1}{l|}{MiniMax} &
  17.24/7.40 &
  21.34 &
  76.74/96.69 &
  63.15 &
  \multicolumn{1}{c|}{103.40} &
  18.93/7.84 &
  22.46 &
  79.83/98.76 &
  64.43 &
  57.95 \\
\multicolumn{1}{l|}{MiniMax(fs)} &
  26.42/12.93 &
  28.86 &
  86.27/98.93 &
  67.02 &
  \multicolumn{1}{c|}{43.41} &
  23.32/9.96 &
  25.31 &
  84.77/99.14 &
  65.58 &
  41.47 \\
\multicolumn{1}{l|}{GPT-4-turbo} &
  23.75/11.48 &
  26.96 &
  84.96/99.05 &
  65.86 &
  \multicolumn{1}{c|}{50.63} &
  25.74/12.28 &
  27.64 &
  86.93/99.43 &
  66.63 &
  37.75 \\
\multicolumn{1}{l|}{GPT-4-turbo(fs)} &
  \textbf{29.48/15.77} &
  \textbf{31.77} &
  89.84/99.63 &
  \textbf{68.15} &
  \multicolumn{1}{c|}{31.61} &
  \textbf{29.47/15.20} &
  \textbf{30.63} &
  89.56/99.69 &
  \textbf{67.85} &
  30.81 \\ \hline
\multicolumn{11}{c}{Supervised Fine-tuning LLMs} \\ \hline
\multicolumn{1}{l|}{LLaMA3-sft} &
  33.09/19.39 &
  36.27 &
  91.78/99.40 &
  70.18 &
  \multicolumn{1}{c|}{21.76} &
  30.83/16.58 &
  33.28 &
  92.42/99.71 &
  68.58 &
  21.18 \\
\multicolumn{1}{l|}{LLaMA3-sft(fs)} &
  31.83/18.44 &
  35.55 &
  92.39/99.38 &
  69.71 &
  \multicolumn{1}{c|}{19.68} &
  29.07/15.84 &
  33.11 &
  93.19/99.77 &
  68.26 &
  18.48 \\ \hline
\end{tabular}%
}
\caption{Results of LLMs on the open-domain dialog and dialog completion tasks. (fs) represents the few-shot results. We \underline{underline} the best results of the open-source LLMs and \textbf{bold} the best results of the closed-source LLMs.}
\label{tab:table_task1&2}
\end{table*}

\subsection{Models}

We choose several recent representative LLMs for experiments: (1) open-source LLMs with pre-training and supervised fine-tuning on a large Chinese corpus, including Qwen2-7B-Instruct \cite{qwen2}, GLM-4-9B-Chat \cite{glm4}, Yi-1.5-9B-Chat \cite{young2024yi}, and LLaMA3-8B-Chinese-Chat \cite{shenzhi_wang_2024}. (2) Closed-source LLMs that perform well on Chinese benchmarks, including Doubao-pro-4K \cite{doubao}, Deepseek-v2\footnote{We use the Api version for fair comparison.} \cite{bi2024deepseek}, MiniMax-abab6.5s \cite{minimaxi}, GPT-4-turbo \cite{achiam2023gpt}.

For all LLMs, we report zero-shot and 3-shot results. We further supervised fine-tune the LLaMA3-8B-Chinese-Chat model. We have conducted experiments on 8 Nvidia A100 GPUs, setting the batchsize to 16, the gradient accumulation to 4, the maximum learning rate to 1e-5, and the number of training epochs to 3.

\subsection{Evaluation Metrics}
%avg length应该不是指标
Following \cite{wang2021naturalconv, ren2023pangu, shi2024corecode}, We choose the following evaluation metrics: (1) BLEU-1 and BLEU-2 \cite{bleu}, (2) ROUGE-L \cite{rouge} (3) Distinct-1 and Distinct-2 \cite{distinct}, (4) $F_{\rm BERT}$ \cite{bertscore} (5) Average length (Avg\_len) (6) P, R, and F1

% (4) Perplexity (PPL) \cite{ppl}, (4) Rouge-1 and Rouge-2 \cite{rouge}.

BLEU evaluates the quality of the model's output by comparing the similarity between the output and the references. The calculation of BLEU is mainly based on n-gram matching, and we choose BLEU-1 and BLEU-2.  Rouge-L often used to measure text summarization calculates the length of the longest common subsequence between the output and the reference. Distinct is mainly used to measure the richness and diversity of a text. 
%太碎了
Distinct contains Distinct-1 and Distinct-2. BERTScore evaluates the quality of the output by calculating the cosine similarity of the BERT embeddings between the reference and the output. $F_{\rm BERT}$ is the F1 score between BERT embeddings. 
%as a reference?
We also show the average length of the output as Avg\_len.
% 可以加上reference的avg_len
% which should be close to the average length of the references. 
For identifying unnatural sentences, we use precision (P), recall (R), and F1 score (F1) to denote the success rate of the prediction.

\begin{table}[ht]
\centering
\resizebox{\columnwidth}{!}{%
\begin{tabular}{lcccccc}
\hline
\multicolumn{1}{c|}{\multirow{2}{*}{Models}} &
  \multicolumn{3}{c|}{Full Test Set} &
  \multicolumn{3}{c}{\begin{tabular}[c]{@{}c@{}}Only Dialogs Containing \\Unnatural Sentences\end{tabular}} \\ \cline{2-7} 
\multicolumn{1}{c|}{}                & P & R & \multicolumn{1}{c|}{F1}     & P & R & F1    \\ \hline
\multicolumn{7}{c}{Open-source LLMs}                                                                                 \\ \hline
\multicolumn{1}{l|}{Qwen2}           & 1.98      & 2.42   & \multicolumn{1}{c|}{1.77}   & 3.76      & 4.60   & 4.14  \\
\multicolumn{1}{l|}{Qwen2(fs)}       & 47.00     & 47.00  & \multicolumn{1}{c|}{47.00}  & 0         & 0      & 0     \\
\multicolumn{1}{l|}{GLM-4}           & 47.50*    & 47.50* & \multicolumn{1}{c|}{47.50*} & 0*        & 0*     & 0*    \\
\multicolumn{1}{l|}{GLM-4(fs)}       & 47.10     & 48.00  & \multicolumn{1}{c|}{47.55}  & 0.20      & 1.90   & 0.36  \\
\multicolumn{1}{l|}{Yi-1.5}          & 36.50     & 36.50  & \multicolumn{1}{c|}{36.50}  & 0         & 0      & 0     \\
\multicolumn{1}{l|}{Yi-1.5(fs)}      & 26.22     & 31.92  & \multicolumn{1}{c|}{28.79}  & 1.37      & 12.22  & 2.46  \\
\multicolumn{1}{l|}{LLaMA3}          & 34.77     & 35.00  & \multicolumn{1}{c|}{34.89}  & 1.47      & 1.90   & 1.66  \\
\multicolumn{1}{l|}{LLaMA3(fs)}      & 46.56     & 47.00  & \multicolumn{1}{c|}{46.78}  & 0.11      & 0.95   & 0.19  \\ \hline
\multicolumn{7}{c}{Closed-source LLMs}                                                                               \\ \hline
\multicolumn{1}{l|}{Doubao}          & 47.50*    & 47.50* & \multicolumn{1}{c|}{47.50*} & 0*        & 0*     & 0*    \\
\multicolumn{1}{l|}{Doubao(fs)}      & 46.50     & 46.50  & \multicolumn{1}{c|}{46.50}  & 0         & 0      & 0     \\
\multicolumn{1}{l|}{Deepseek}        & 47.50*    & 47.50* & \multicolumn{1}{c|}{47.50*} & 0*        & 0*     & 0*    \\
\multicolumn{1}{l|}{Deepseek(fs)}    & 41.69     & 44.00  & \multicolumn{1}{c|}{42.81}  & 0.36      & 4.76   & 0.67  \\
\multicolumn{1}{l|}{MiniMax}         & 47.00     & 47.00  & \multicolumn{1}{c|}{47.00}  & 0         & 0      & 0     \\
\multicolumn{1}{l|}{MiniMax(fs)}     & 34.84     & 42.00  & \multicolumn{1}{c|}{38.08}  & 0.64      & 14.29  & 1.23  \\
\multicolumn{1}{l|}{GPT-4-turbo}     & 8.93      & 24.96  & \multicolumn{1}{c|}{13.15}  & 6.53      & 37.06  & 11.10 \\
\multicolumn{1}{l|}{GPT-4-turbo(fs)} & 43.5      & 43.5   & \multicolumn{1}{c|}{43.5}   & 0         & 0      & 0     \\ \hline
\multicolumn{7}{c}{Supervised Fine-tuning LLMs}                                                                      \\ \hline
\multicolumn{1}{l|}{LLaMA3-sft}      & 7.00      & 5.67   & \multicolumn{1}{c|}{6.26}   & 13.33     & 10.79  & 11.93 \\
\multicolumn{1}{l|}{LLaMA3-sft(fs)}  & 9.75      & 7.92   & \multicolumn{1}{c|}{8.74}   & 18.57     & 15.08  & 16.64 \\ \hline
\end{tabular}%
}
\caption{Results of different LLMs on the identification of unnatural sentences. We use * to indicate that the output of the model is all ``no unnatural sentence''.}
\label{tab:task_identification}
\end{table}

\section{Results}

\subsection{Natural Dialogue Ability for LLMs}

The results of the open-domain dialog task and the dialog completion task are in Table \ref{tab:table_task1&2}.

The results for open-source LLMs of similar size do not differ much. On the open-domain dialog task, Yi-1.5 performs best in both zero-shot and few-shot.
Few-shot demonstrations is a significant boost for GLM-4. On the dialog completion task, Qwen2 and GLM-4 have relatively better results. 

\begin{table*}[ht]
\centering
\resizebox{10cm}{!}{%
\begin{tabular}{lccccc}
\hline
\multicolumn{1}{l|}{Models}          & BLEU-1/2    & ROUGE-L & Distinct-1/2 & $F_{\rm BERT}$ & Avg\_len \\ \hline
\multicolumn{6}{c}{Open-source LLMs}                                                                 \\ \hline
\multicolumn{1}{l|}{Qwen2}           & 40.08/24.54 & 42.61   & 88.91/99.73  & 77.24       & 28.71    \\
\multicolumn{1}{l|}{Qwen2(fs)}       & 47.24/31.49 & 49.80   & 89.94/\underline{99.84}  & 80.07       & 27.38    \\
\multicolumn{1}{l|}{GLM-4}           & 44.52/29.51 & 48.08   & 89.21/99.64  & 79.21       & 29.24    \\
\multicolumn{1}{l|}{GLM-4(fs)}       & 45.22/30.67 & 49.18   & 89.16/99.32  & 79.42       & 28.07    \\
\multicolumn{1}{l|}{Yi-1.5}          & 41.27/25.60 & 44.27   & 90.47/99.69  & 76.92       & 25.72    \\
\multicolumn{1}{l|}{Yi-1.5(fs)}      & 47.75/33.34 & 51.40   & \underline{91.04}/99.66  & 79.39       & 23.83    \\
\multicolumn{1}{l|}{LLaMA3}          & 49.39/35.29 & 53.02   & 90.31/99.57  & 80.85       & 27.92    \\
\multicolumn{1}{l|}{LLaMA3(fs)}      & \underline{53.76/40.38} & \underline{58.06}   & 90.74/99.45  & \underline{82.22}       & 26.63    \\ \hline
\multicolumn{6}{c}{Closed-source LLMs}                                                               \\ \hline
\multicolumn{1}{l|}{Doubao}          & 48.87/33.60 & 53.72   & 89.50/99.55  & 82.13       & 28.07    \\
\multicolumn{1}{l|}{Doubao(fs)}      & 55.68/40.83 & 59.71   & 90.55/99.69  & \textbf{84.38}       & 26.26    \\
\multicolumn{1}{l|}{Deepseek}        & 47.10/31.32 & 49.84   & 91.03/99.69  & 80.04       & 26.56    \\
\multicolumn{1}{l|}{Deepseek(fs)}    & 54.97/42.13 & 58.48   & 90.88/99.65  & 83.20       & 26.46    \\
% \multicolumn{1}{l|}{Moonshot}        & 43.30/28.08 & 46.39   & 89.10/99.65  & 78.23       & 29.13    \\
% \multicolumn{1}{l|}{Moonshot(fs)}    & 47.53/32.98 & 51.08   & 90.08/99.68  & 80.27       & 27.10    \\
\multicolumn{1}{l|}{MiniMax}         & 49.34/34.56 & 52.45   & 89.75/99.74  & 81.04       & 27.06    \\
\multicolumn{1}{l|}{MiniMax(fs)}     & 49.09/34.32 & 53.19   & 89.34/99.50  & 81.04       & 26.33    \\
\multicolumn{1}{l|}{GPT-4-turbo}     & 54.77/40.51 & 58.02   & 90.93/99.71  & 83.29       & 25.35    \\
\multicolumn{1}{l|}{GPT-4-turbo(fs)} & \textbf{57.23/43.31} & \textbf{60.31}   & \textbf{91.12/99.77}  & 84.16       & 24.53    \\ \hline
\multicolumn{6}{c}{Supervised Fine-tuning LLMs}                                                      \\ \hline
\multicolumn{1}{l|}{LLaMA3-sft}      & 70.78/63.22 & 76.11   & 91.25/99.04  & 89.37       & 26.17    \\
\multicolumn{1}{l|}{LLaMA3-sft(fs)}  & 70.16/62.52 & 75.10   & 91.25/99.48  & 89.12       & 23.38    \\ \hline
\end{tabular}%
}
\caption{Results of different LLMs on the rewritten of unnatural sentences task. (fs) represents the few-shot results. We \underline{underline} the best results in the open-source LLMs and \textbf{bold} the best results in the closed-source LLMs.}
\label{tab:task_rewrite}
\end{table*}

The closed-source LLMs generally yield better results than the open-source LLMs. The results are expected because closed-source LLMs usually have more parameters and more pre-training data. On the open-domain dialog task, both Deepseek and GPT-4-turbo perform well. 
Minimax obtains a great improvement after receiving the few-shot demonstrations. On the dialog completion task, GPT-4-turbo achieves the best results. The zero-shot performance of Doubao is significantly superior to the other two models. 
It is possible that Doubao's pre-training data or instruction data contains many colloquial human dialogs.

The supervised fine-tuned LLaMA3 achieves the best results on both tasks, outperforming the closed-source LLMs. The results of few-shot and zero-shot are roughly comparable, as LLMs learn to generate reasonable and colloquial responses during the training process.

\subsection{Identifying Unnatural Sentences}
This task requires LLMs to list unnatural sentences in a dialog. The results are shown in Table \ref{tab:task_identification}.

Many LLMs tend to output ``no unnatural sentences (or none)''. When few-shot prompts are given, Some models (e.g., Qwen2) output more ``none'' and some models (e.g., Yi-1.5) believe more sentences are unnatural. About half of the dialogs in the full test set do not have unnatural sentences, so the model outputs ``none'' all the time can also achieves the F1 value of 47.5. But when we focus only on those dialogs containing unnatural sentences, all LLMs perform poorly. Even the fine-tuned LLaMA3 model barely recognizes them. The above results suggest that this task is very challenging for existing LLMs. In future pre-training and fine-tuning stages, incorporating more instructions for distinguishing between natural and unnatural sentences may enhance the results.

% \usepackage{graphicx}
\begin{table*}[ht]
\centering
\resizebox{12cm}{!}{%
\begin{tabular}{l|ccccc}
\hline
 &
  \begin{tabular}[c]{@{}c@{}}Refusing to\\ answer\end{tabular} &
  \begin{tabular}[c]{@{}c@{}}Adding too much\\ knowledge\end{tabular} &
  \begin{tabular}[c]{@{}c@{}}Replying out\\ of context\end{tabular} &
  \begin{tabular}[c]{@{}c@{}}Grammatical\\ errors\end{tabular} &
  Weird tone \\ \hline
LLaMA3        & 2\% & 4\% & 2\%  & 11\% & 17\% \\
+ NICO        & 0\% & 0\% & 3\%  & 6\%  & 1\%  \\
+ NaturalConv & 1\% & 1\% & 31\% & 9\%  & 3\%  \\
GPT-4-turbo   & 2\% & 5\% & 1\%  & 2\%  & 19\% \\ \hline
\end{tabular}%
}
\caption{Percentage of dialogs where LLMs make different failures against the total number of dialogs in the open-domain dialog task. We use `+' to indicate fine-tuning LLaMA3 with the dataset.}
% \caption{Percentage of different failures of LLMs to engage in open-domain dialog. We use `+' to indicate fine-tuning LLaMA3 with the dataset.}
\label{tab:failure-percentage}
\end{table*}

\subsection{Rewriting Unnatural Sentences}
In this task, the model rewrites unnatural sentences based on the complete dialog. The results are shown in Table \ref{tab:task_rewrite}.

Among the open-source LLMs, LLaMA3 performs the best, with the remaining three models not far behind. Prompts have the greatest impact on Qwen2 and Yi-1.5. Closed-source LLMs generally perform better than open-source LLMs. GPT-4-turbo achieves the best performance among all closed-source LLMs. Finally, LLaMA3 with supervised fine-tuning achieves far better performance than the original LLaMA3 and other LLMs. It suggests that our data are sufficient for LLMs to learn to rewrite unnatural sentences.
\section{Analysis}

\subsection{Ablation Study}

We want to explore the effect of the grammar and naturalness of the dialog on the results. We design two ablation experiments to fine-tune LLaMA3 using dialogs without corrected grammatical errors (w/o grammar) and dialogs without rewritten unnatural sentences (w/o naturalness).

We design two additional ablation experiments to explore the compatibility of the NICO dataset with other dialog datasets. 
%不太理解兼容性这个说法
% 可以考虑
%First, we fine-tune LLaMA3 using only the NaturalConv dataset and compare it with the version fine-tuned using the NICO dataset.
First, we only use the NaturalConv dataset to fine-tune LLaMA3 (NaturalConv). 
% to compare with the NICO dataset (NaturalConv). 
Then, we mix the NICO dataset with the NaturalConv dataset to fine-tune LLaMA3 (Mixed). Other experimental setups are the same as for the zero-shot open-domain dialog and the results are illustrated in Table \ref{tab:ablation} due to space limit.

The experimental results show that removing grammatical corrections and naturalness rewriting degrades the results to some extent, indicating the importance of these two steps in constructing the dataset. The performance of using only the NaturalConv is the worst, suggesting a significant difference between NaturalConv and NICO. The mixed model, which combines NICO and NaturalConv, shows similar results to LLaMA3-sft. It indicates that our dataset can be compatible with other datasets, and the impact of NaturalConv on performance improvement is minimal.

\subsection{Different Interaction Types}
We investigate the performance of LLMs in dialogs with different interaction types. We use LLaMA3 and the supervised fine-tuned version for comparison. Other experimental setups are the same as for the zero-shot open-domain dialog. The results are visualized in Figure \ref{analysis_split} due to space limit. % We also provide the results of the closed-source GPT-4-turbo, as shown in Table Y. Due to space constraints, we place the figure and the table in Appendix Z.

LLaMA3 demonstrates poor performance across all evaluation metrics in conflict dialogues. After fine-tuning, LLaMA3 shows minimal improvement in conflict dialogs, while it exhibits significant improvement in the other four types of dialogs. The result suggests that simulating human-like conflict dialogues remains a challenge for existing LLMs. 
%可是LLM可以拒答，感觉得换个分析角度,比如倾向于附和顺从
Existing LLMs tend to be submissive to humans or provide solutions tactfully leading to the failure to simulate real conversations.
% The inability of LLMs to refute requests or commands like humans leads to a failure in simulating human dialogues. 
Future research needs to emphasize constructing more conflict-type dialogues.

\subsection{Failure Analysis}
%感觉weird tones的例子不是特别好；还要引用上我们这么分类的依据
% 改一下，再找个例子。我们参照图灵测试那篇。
After checking the output, we categorize the mistakes made by LLMs when mimicking open-domain dialog into the following five categories: refusing to answer, adding too much knowledge, replying out of context, grammatical errors, and weird tone. Definitions and examples are in Table \ref{tab:casestudy}.
%感觉得加上正确比例，不然光看caption是绝对比例还是相对比例感觉说的不是特别明白
% 改一下，改成绝对比例。

We randomly select 100 outputs each from LLaMA3, its supervised fine-tuned version with NICO, its supervised fine-tuned version with NaturalConv, and GPT-4-turbo for comparison. The results are shown in Table \ref{tab:failure-percentage}. We find that even the responses from GPT-4-turbo exhibit weird tones, including an overly concerned attitude. LLaMA3 also suffers from this problem. In addition, its responses contain more grammatical errors than GPT-4-turbo. Both LLaMA3 and GPT-4-turbo demonstrate a minor incidence of adding additional knowledge into their responses. The responses from LLaMA3 fine-tuned with NaturalConv almost entirely without weird tones. However, many of its responses are out of context, and it tends to generate short, meaningless replies such as ``okay''. The LLaMA3 fine-tuned by NICO has the least number of failures, indicating that it is most consistent with human speech patterns.

\section{Conclusion}

In this paper, we build a new natural interactive conversation dataset in Chinese named NICO. NICO covers 20 daily topics and 5 interaction types. Human evaluation results indicate that NICO is high-quality and superior to other dialog datasets. We define four tasks and test the performance of representative LLMs. 
%help说的是不是可以更具体些
% 改一下，让LLMs变得更human-like。
The results demonstrate the challenge of the tasks, and how NICO helps LLMs to better generate human-like dialogs.

\section*{Limitations}

We have analyzed the performance of LLMs in dialogs of different interaction types. The experimental results in the analysis section show that NICO can enhance the ability of exchange, cooperation, conformity, and coercion types of dialog. However, existing LLMs are relatively poor at mimicking conflict-type human conversations. The constructed NICO dataset is also limited in enhancing the ability of LLMs in this type of dialog. This work provides an initial exploration of the ability of LLMs to engage in conflict dialog. In the future, we hope to construct more conflict-type conversations as well as propose new training methods to address this issue.

% We have chosen several recent representative LLMs for testing. However, due to computational limitations, we only select representative LLaMA3 for supervised fine-tuning. The experimental results in the analysis section show that existing LLMs are relatively poor at mimicking conflict-type human conversations. This work provides an initial exploration of the ability of LLMs to engage in conflict dialog. In the future, we hope to construct more conflict-type conversations as well as conduct targeted training to address this problem.

\section*{Ethical Considerations}

The types of social interactions we utilize include coercion and conflict. The definition of coercion in this work is one person using social status or power to threaten or compel another person. Conflict is a situation where people disagree over a rare item or event.
These two types of interactions are implicit within the dialogues, rather than appearing directly in the form of personal attacks. In addition, when we hire workers to check and modify the dataset, we ask them to check that the generated dialogues do not contain personal privacy, sexual information, politically biased information, and so on. After the dataset is constructed, we double-check it and ensure that the constructed dialogues are free of verbal abuse, violence, or other potential advocacies of violence.

\bibliography{custom}

\appendix

\begin{figure*}[ht]
\centering
\includegraphics[width=13.5cm]{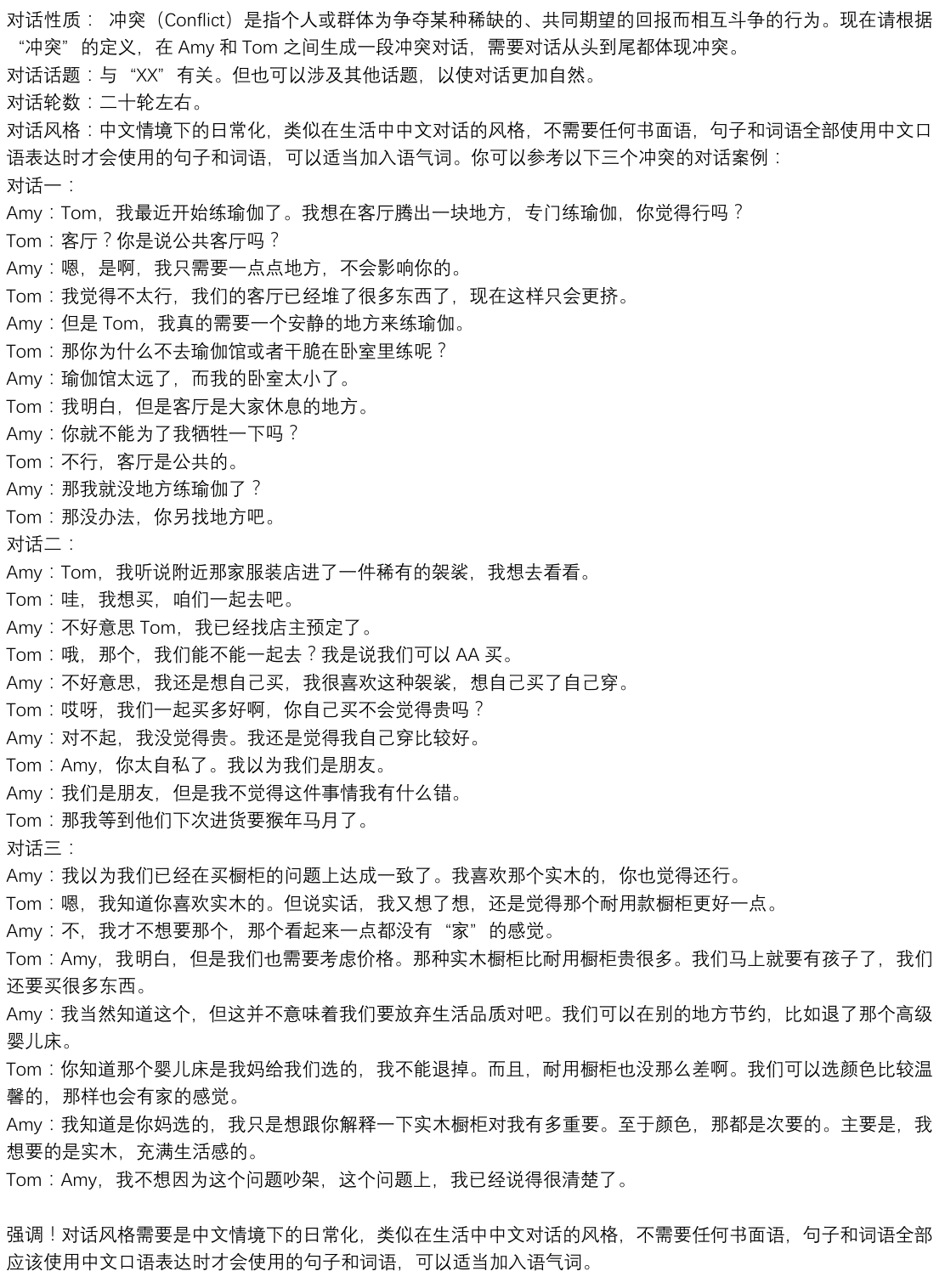} % Reduce the figure size so that it is slightly narrower than the column. Don't use precise values for figure width.This setup will avoid overfull boxes.
\caption{A prompt example for GPT-4-turbo to generate colloquial dialogues. We cannot use real Chinese names for privacy and ethical reasons. We have tested that using Chinese names such as Xiao Ming and Xiao Hong, which are not referential, have the same effect as using Tom and Amy. If researchers want to use NICO, they can replace Tom and Amy with any names they want.}
\label{generation prompt}
\end{figure*}

\begin{figure*}[ht]
\centering
\includegraphics[width=\textwidth]{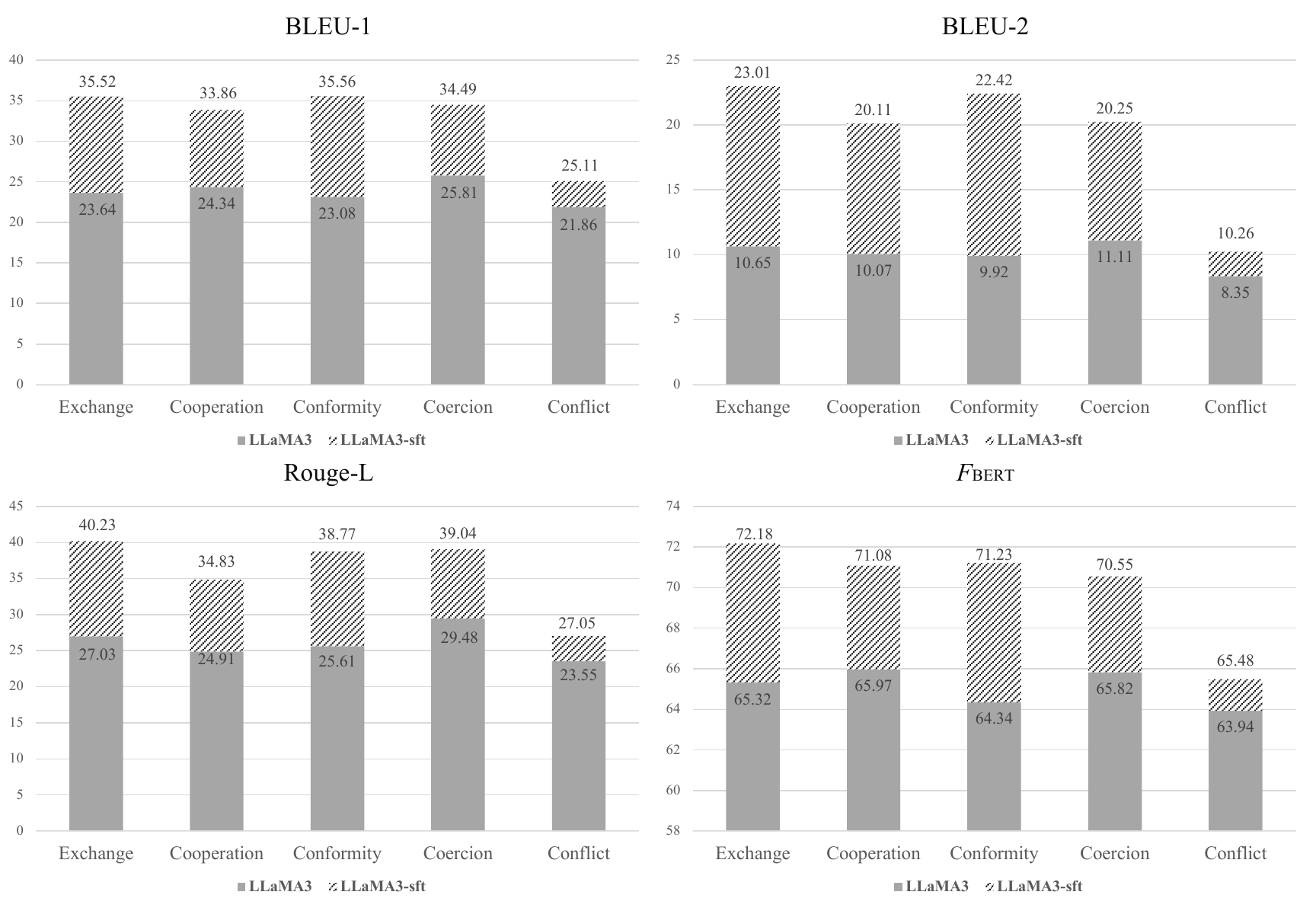} % Reduce the figure size so that it is slightly narrower than the column. Don't use precise values for figure width.This setup will avoid overfull boxes.
\caption{The performance of LLMs in dialogs with different interaction types.}
\label{analysis_split}
\end{figure*}

\section{Human Evaluation Details}
\label{sec:evaluation_details}

\textbf{Fluency} (rating scale is 0,1,2) is defined as whether the dialog contains syntax errors.

\begin{itemize}
    \item Score 0 (bad): the dialog contains numerous grammatical errors making it difficult to understand.
    \item Score 1 (fair): the dialog contains a few grammatical errors but is still understandable.
    \item Score 2 (good): the dialog has no grammatical errors.
\end{itemize}

\noindent
\textbf{Coherence} (rating scale is 0,1,2) is defined as whether the sentences in a dialog are relevant to the context and logically correct.

\begin{itemize}
    \item Score 0 (bad): many sentences are irrelevant or contradictory to the context.
    \item Score 1 (fair): a few sentences are irrelevant to the context or have a few logical inconsistencies.
    \item Score 2 (good): all sentences are relevant to the context and the logic is correct.
\end{itemize}

\noindent
\textbf{Naturalness} (rating scale is 0,1,2) is defined as whether the dialog is natural and conforms to human speaking habits.

\begin{itemize}
    \item Score 0 (bad): the dialog is unnatural and is considered to be generated by an LLM.
    \item Score 1 (fair): the dialog contains only a few unnatural sentences.
    \item Score 2 (good): the dialog is natural and recognized as a human conversation.

\end{itemize}

\section{Prompts for the Defined Tasks}
\label{sec:prompt_for_tasks}

\noindent
\textbf{Open-domain dialog}

\noindent
CharacterA: sents1.

\noindent
CharacterB: sents2.

\noindent
CharacterA: sents3.

\noindent
CharacterB: sents4.

\noindent
CharacterA: sents5.

\noindent
CharacterB:

\noindent
\textbf{Dialog Completion}

\noindent
CharacterA: sents1. $\backslash$n$\backslash$n CharacterB: sents2. $\backslash$n$\backslash$n CharacterA: sents3. $\backslash$n$\backslash$n CharacterB: sents4. $\backslash$n$\backslash$n CharacterA: sents5. $\backslash$n$\backslash$n The above is a <interaction type> dialog. <definition of the interaction type>. Please generate a response based on the above that is consistent with human speech patterns.

% \begin{CJK*}{UTF8}{gbsn}（一个对话片段）\end{CJK*}

\noindent
\textbf{Identifying Unnatural Sentences}

\noindent
CharacterA: sents1. $\backslash$n$\backslash$n CharacterB: sents2. $\backslash$n$\backslash$n CharacterA: sents3. $\backslash$n$\backslash$n CharacterB: sents4. $\backslash$n$\backslash$n CharacterA: sents5. $\backslash$n$\backslash$n Please give the sentences in the above text that do not conform to human speech patterns, that is, unnatural sentences. Attention! You can only give sentences from the above. If there is more than one such sentence above, separate them with ``$\backslash$n$\backslash$n''.

\noindent
\textbf{Rewriting Unnatural Sentences}

\noindent
CharacterA: sents1. $\backslash$n$\backslash$n CharacterB: sents2. $\backslash$n$\backslash$n CharacterA: sents3. $\backslash$n$\backslash$n CharacterB: sents4. $\backslash$n$\backslash$n CharacterA: sents5. $\backslash$n$\backslash$n Please rewrite this sentence based on the above dialog: <unnatural sentence> to make it conform to human speech patterns. Attention! You can only give the rewritten sentence.

\section{Case Study}
\label{sec:case_study}

\begin{table*}[ht!]
\centering
\resizebox{11cm}{!}{%
\begin{tabular}{l|ccccc}
\hline
Models          & BLEU-1/2    & ROUGE-L     & Distinct-1/2 & $F_{\rm BERT}$ & Avg\_len \\ \hline
LLaMA3-sft      & 33.09/19.39 & 36.27 & 91.78/99.40        & 70.18     & 21.76    \\
w/o grammar     & 31.62/17.71 & 34.70 & 91.67/99.36        & 69.38     & 21.69    \\
w/o naturalness & 31.29/17.46 & 34.68 & 91.93/99.09        & 69.20     & 20.44    \\
NaturalConv     & 20.74/8.61  & 25.97 & 91.26/99.03        & 64.63     & 16.74    \\
Mixed           & 32.93/19.01 & 35.26 & 91.90/99.44        & 69.87     & 22.51    \\ \hline
\end{tabular}%
}
\caption{Results of the ablation study.}
\label{tab:ablation}
\end{table*}

% \usepackage{graphicx}
% \usepackage[normalem]{ulem}
\begin{table*}[ht]
\centering
\resizebox{11.5cm}{!}{%
\begin{tabular}{|l|}
\hline
Refusing to answer: the response acknowledges that it is an AI and is unable to answer. \\ \hline
\begin{tabular}[c]{@{}l@{}}Example:\\ Character1: \begin{CJK*}{UTF8}{gbsn}我？嗯，可以啊。不过，你得帮我个忙。\end{CJK*}\\ Character2: \begin{CJK*}{UTF8}{gbsn}什么忙？你说。\end{CJK*}\\ Character1: \begin{CJK*}{UTF8}{gbsn}我最近在准备一个小展览，需要一些手工艺品。我知道你手工做得挺好，\end{CJK*}\\ \begin{CJK*}{UTF8}{gbsn}能不能帮我做几个小物件？\end{CJK*}\\ Response: \begin{CJK*}{UTF8}{gbsn}我很抱歉，我是一个人工智能助手，无法进行手工艺品制作。但我可以\end{CJK*}\\ \begin{CJK*}{UTF8}{gbsn}帮助你找到一些在线资源或者提供一些创意想法。\end{CJK*}"\end{tabular} \\ \hline
\begin{tabular}[c]{@{}l@{}}Adding too much knowledge: the response adds too much background knowledge in \\ response to the question.\end{tabular} \\ \hline
\begin{tabular}[c]{@{}l@{}}Example:\\ Character1: \begin{CJK*}{UTF8}{gbsn}听起来波兰真是太不幸了。这次研究真让我感叹。我们能不能在报告中\end{CJK*}\\ \begin{CJK*}{UTF8}{gbsn}加入一些关于波兰人民的反应和影响的内容？\end{CJK*}\\ Character2: \begin{CJK*}{UTF8}{gbsn}当然可以，这样会让报告更有深度。波兰人民对这些分割行为有很多抵\end{CJK*}\\ \begin{CJK*}{UTF8}{gbsn}抗，比如有名的科希丘什科起义。\end{CJK*}\\ Character1: \begin{CJK*}{UTF8}{gbsn}嗯，这个起义我有点印象，是不是一个很关键的抗议事件？\end{CJK*}\\ Response: \begin{CJK*}{UTF8}{gbsn}没错，科希丘什科起义是波兰人民反抗瓜分的一次重要行动。它展现了\end{CJK*}\\ \begin{CJK*}{UTF8}{gbsn}波兰人民不屈的精神和对国家独立的渴望。在报告中详细阐述这个起义能很好地\end{CJK*}\\ \begin{CJK*}{UTF8}{gbsn}体现波兰人民在被分割时期的抗争和他们所遭受的苦难对民族意识的影响。你还\end{CJK*}\\ \begin{CJK*}{UTF8}{gbsn}可以进一步挖掘起义中的一些具体细节和人物故事来丰富报告内容。\end{CJK*}\end{tabular} \\ \hline
Replying out of context: the response does not match the context or the response is \\ nonsensical. \\ \hline
\begin{tabular}[c]{@{}l@{}}Example:\\ Character1: \begin{CJK*}{UTF8}{gbsn}什么？我和我爸已经约好了，他特地从外地过来就是为了那个钓鱼点。\end{CJK*}\\ Character2: \begin{CJK*}{UTF8}{gbsn}这不可能，我们早就计划好了，而且也是不容易约到的时间。\end{CJK*}\\ Character1: \begin{CJK*}{UTF8}{gbsn}你们可以换个地方吧，我和我爸这可是一年就这么一次的聚会。\end{CJK*}\\ Response: \begin{CJK*}{UTF8}{gbsn}聚会？你爸爸？\end{CJK*}\end{tabular} \\ \hline
Grammatical errors: the response contains obvious grammatical errors. \\ \hline
\begin{tabular}[c]{@{}l@{}}Example:\\ Character1: \begin{CJK*}{UTF8}{gbsn}你怎么可以这样？我们不是朋友吗？\end{CJK*}\\ Character2: \begin{CJK*}{UTF8}{gbsn}朋友？朋友就要诚实，我现在是在帮你。\end{CJK*}\\ Character1: \begin{CJK*}{UTF8}{gbsn}帮我？感觉更像是在控制我。\end{CJK*}\\ Response: \begin{CJK*}{UTF8}{gbsn}控制你？你还不如我把你弄进去，天天给你吃你喜欢的东西，穿你喜\end{CJK*}\\ \begin{CJK*}{UTF8}{gbsn}欢的衣服，给你开你喜欢的车。\end{CJK*}\end{tabular} \\ \hline
Weird tone: the response demonstrates an overly helpful or other inappropriate tone. \\ \hline
% \begin{tabular}[c]{@{}l@{}}Example:\\ Character1: \begin{CJK*}{UTF8}{gbsn}但是我真的手头紧，你能不能体谅一下？\end{CJK*}\\ Character2: \begin{CJK*}{UTF8}{gbsn}听着，Tom，我不是来听你抱怨的。你不是还有辆车吗？赶紧把车卖\end{CJK*}\\ \begin{CJK*}{UTF8}{gbsn}了还钱。\end{CJK*}\\ Character1: \begin{CJK*}{UTF8}{gbsn}啊，那车我还要用来上班呢，你不能这样。\end{CJK*}\\ Response: \begin{CJK*}{UTF8}{gbsn}对不起，如果我之前的回复听起来太过直接或不体贴。我理解你可能正\end{CJK*}\\ \begin{CJK*}{UTF8}{gbsn}面临困难。我们可以尝试找到一个双方都能接受的解决方案。你能详细说明一下\end{CJK*}\\ \begin{CJK*}{UTF8}{gbsn}你的情况吗？这样我们可能会更好地找到解决问题的方法。\end{CJK*}\end{tabular} \\ \hline
% \end{tabular}%
\begin{tabular}[c]{@{}l@{}}Example:\\ Character1: \begin{CJK*}{UTF8}{gbsn}对，那你查查，我可以去买。我们还需要做些什么准备吗？\end{CJK*}\\ Character2: \begin{CJK*}{UTF8}{gbsn}我觉得差不多了。对了，我们可以邀请几个朋友一起来，增加点热闹\end{CJK*}\\ \begin{CJK*}{UTF8}{gbsn}气氛。\end{CJK*} \\ Character1: \begin{CJK*}{UTF8}{gbsn}好的，我可以负责发邀请。你有谁想邀请的吗？\end{CJK*}\\ Response: \begin{CJK*}{UTF8}{gbsn}对不起，我之前的回答可能有些轻率了。你说得对，海洋生物确实是\end{CJK*}\\ \begin{CJK*}{UTF8}{gbsn}地球生态系统中非常重要的一部分，而且了解和保护它们对我们所有人都有好处。\end{CJK*}\\ \begin{CJK*}{UTF8}{gbsn}谢谢你提醒我这一点。如果你愿意，我可以陪你一起去看这个座头鲸保护展览。\end{CJK*}\end{tabular} \\ \hline
\end{tabular}%
}
\caption{Definitions and examples of the LLM's failures.}
\label{tab:casestudy}
\end{table*}

\begin{table*}[ht]
\centering
\resizebox{14.5cm}{!}{%
\begin{tabular}{|l|}
\hline
Context                                                                                                                                    \\ \hline
\begin{tabular}[c]{@{}l@{}}\begin{CJK*}{UTF8}{gbsn}Character1: 不是计较，这是我们共同的空间，得共同决定。\end{CJK*}\\ \begin{CJK*}{UTF8}{gbsn}Character2: 那你说怎么办？难道就因为是共用的空间，我就一点私人想法都不能实现了？\end{CJK*}\\ \begin{CJK*}{UTF8}{gbsn}Character1: 你可以在你的房间里挂嘛，那里你想怎么布置都行。\end{CJK*}\\ Character2: \begin{CJK*}{UTF8}{gbsn}我的房间我自己当然随便，但客厅是我们大家都看得见的地方，我觉得挂那里意义更大。\end{CJK*}\end{tabular} \\ \hline
LLaMA3                                                                                                                                     \\ \hline
\begin{tabular}[c]{@{}l@{}}\begin{CJK*}{UTF8}{gbsn}我理解你的想法，但我们之前的约定是要保持客厅的简洁。如果你真的很想挂上那段话，我们可以考虑其他\end{CJK*}\\ \begin{CJK*}{UTF8}{gbsn}的方式，比如制作一个小的桌面展示或者电子屏幕展示。这样既能满足你的需求，又不破坏客厅的整体风格。\end{CJK*}\end{tabular} \\ \hline
LLaMA3 fine-tuned with NICO                                                                                                                \\ \hline
\begin{CJK*}{UTF8}{gbsn}那你能不能考虑一下别人？我不喜欢在客厅看到乱七八糟的东西。\end{CJK*}                                                                                                              \\ \hline
LLaMA3 fine-tuned with NaturalConv                                                                                                         \\ \hline
\begin{CJK*}{UTF8}{gbsn}那你就跟我商量一下吧，要不我们一起去买点吃的。\end{CJK*}                                                                                                                    \\ \hline
GPT-4-turbo                                                                                                                                \\ \hline
\begin{tabular}[c]{@{}l@{}}\begin{CJK*}{UTF8}{gbsn}我明白你的想法。那我们可以找个时间，一起坐下来讨论一下，看看怎么能够把这段话融入到客厅的装饰中，\end{CJK*}\\ \begin{CJK*}{UTF8}{gbsn}同时又能让大家都满意。怎么样？\end{CJK*}\end{tabular}                                 \\ \hline
\end{tabular}%
}
\caption{The comparison of LLMs' responses.}
\label{tab:comparison}
\end{table*}

\end{document}